\documentclass[lettersize,journal]{IEEEtran}
\usepackage{amsmath,amsfonts}
\usepackage{algorithmic}
\usepackage{algorithm}
\usepackage{array}
\usepackage[caption=false,font=normalsize,labelfont=sf,textfont=sf]{subfig}
\usepackage{textcomp}
\usepackage{stfloats}
\usepackage{url}
\usepackage{verbatim}
\usepackage{graphicx}
\usepackage{cite}
\usepackage{multirow, booktabs, graphicx}
\usepackage{threeparttable}
\begin{document}

\title{DATR: Diffusion-based 3D Apple Tree Reconstruction Framework with Sparse-View}

\author{Tian Qiu,~\IEEEmembership{Graduate Student Member, IEEE}, Alan Zoubi, Yiyuan Lin, Ruiming Du, Lailiang Cheng, Yu Jiang, ~\IEEEmembership{Member, IEEE}
\thanks{Tian Qiu and Yiyuan Lin is with School of Electrical and Computer Engineering, Cornell University, Ithaca, USA
{\tt\small tq42@cornell.edu}}%
\thanks{Alan Zoubi is with Sibley School of Mechanical and Aerospace Engineering, Cornell University, Ithaca, USA}
\thanks{Ruiming Du is with School of Biological and Environmental Engineering, Cornell University, Ithaca, USA}%
\thanks{Lailiang Cheng is with School of Integrative Plant Science, Cornell University, Ithaca, USA}%
\thanks{Yu Jiang is with School of Integrative Plant Science, Cornell University, Geneva, USA
{\tt\small yujiang@cornell.edu}}

}



\maketitle

\begin{abstract}
Digital twin applications offered transformative potential by enabling real-time monitoring and robotic simulation through accurate virtual replicas of physical assets. The key to these systems is 3D reconstruction with high geometrical fidelity. However, existing methods struggled under field conditions, especially with sparse and occluded views. This study developed a two-stage framework (DATR) for the reconstruction of apple trees from sparse views. The first stage leverages onboard sensors and foundation models to semi-automatically generate tree masks from complex field images. Tree masks are used to filter out background information in multi-modal data for the single-image-to-3D reconstruction at the second stage. This stage consists of a diffusion model and a large reconstruction model for respective multi view and implicit neural field generation. The training of the diffusion model and LRM was achieved by using realistic synthetic apple trees generated by a Real2Sim data generator. The framework was evaluated on both field and synthetic datasets. The field dataset includes six apple trees with field-measured ground truth, while the synthetic dataset featured structurally diverse trees. Evaluation results showed that our DATR framework outperformed existing 3D reconstruction methods across both datasets and achieved domain-trait estimation comparable to industrial-grade stationary laser scanners while improving the throughput by $\sim$360 times, demonstrating strong potential for scalable agricultural digital twin systems.

\end{abstract}


\section{Introduction}
\IEEEPARstart{A}gricultural robots are anticipated to address persistent labor challenges in specialty crop production. However, their development is significantly constrained by seasonal crop growth cycles, unlike general-purpose robots. This seasonality limits testing opportunities, extends development timelines, and slows innovation. Digital twin technology offers a promising solution by enabling year-round virtual testing and validation, potentially accelerating development while reducing dependency on seasonal field trials.

High-fidelity 3D models are critical to the effectiveness of digital twins in robotic development, particularly in agricultural applications where robots must interact closely with their environment for both perception and actuation tasks. In controlled environment agriculture, the easy access to crop plants makes high-throughput 3D scanning both accurate and feasible \cite{NPEC2024}. However, reconstructing plants in open fields presents significant challenges because of varying lighting conditions, cluttered background, and frequent occlusions among neighboring plants. Previous studies have investigated both 2D and 3D imaging techniques to overcome these issues. High-resolution terrestrial laser scanners (TLS) offer rich 3D data and are stable to ambient lighting variation \cite{Sun2021, Qiu2024-AppleQSM}, but they are not widely adopted in large agricultural settings due to high sensor cost, power consumption, and limited scalability. In contrast, 2D imaging systems, typically mounted on mobile platforms, are more affordable and adaptable \cite{Liu2018, Majeed2020, Fu2020}, enabling field-scale data collection. Yet their limited views fundamentally restricts 3D reconstruction accuracy, resulting in incomplete geometry due to occlusions, narrow viewing angles, and noisy depth estimation under field conditions.

\begin{figure}[!t]
    \includegraphics[width=\linewidth]{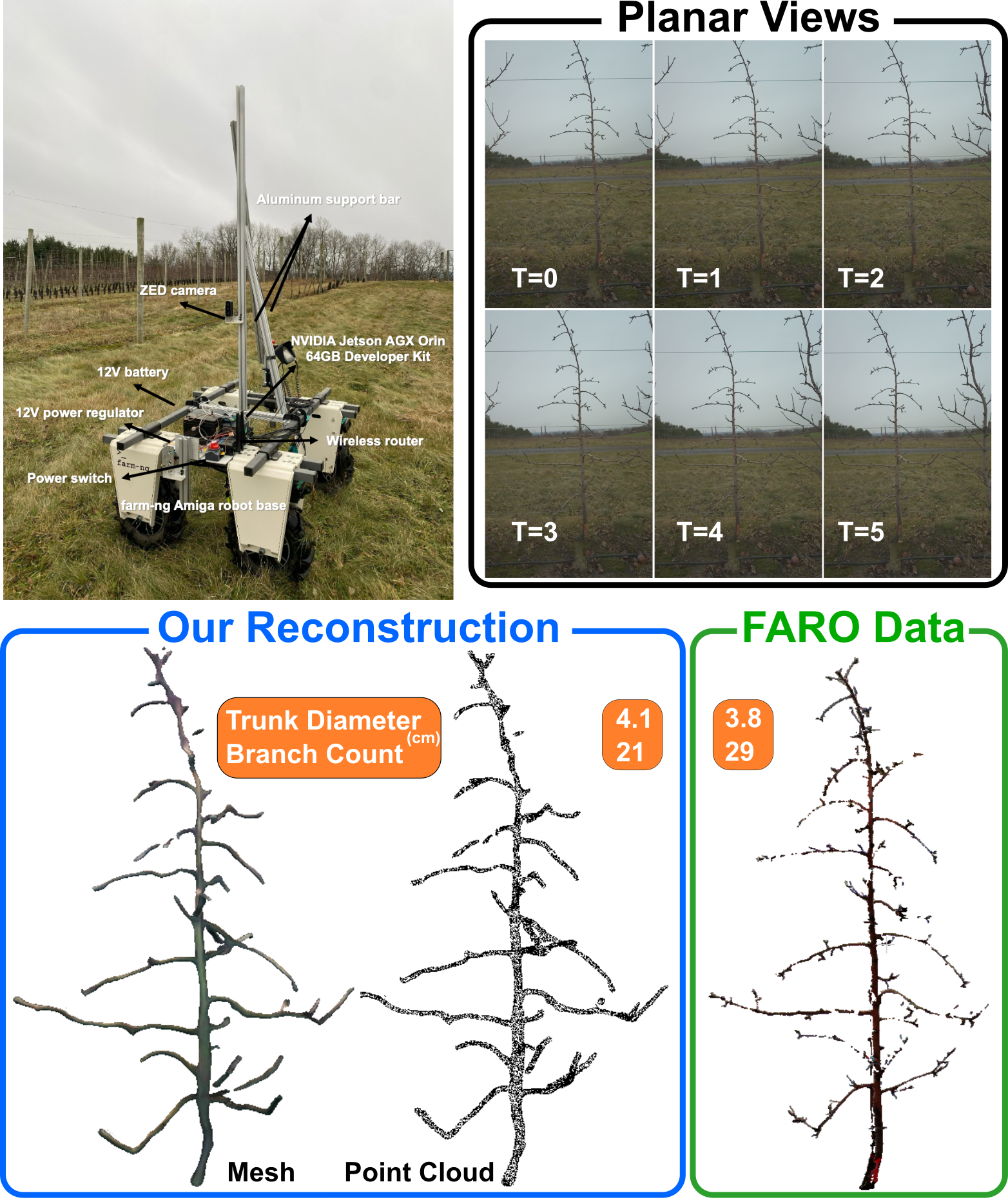}
    \caption{Top: Our AATBot and field tree RGB images. Bottom: single-image-to-3D reconstruction results from our DATR framework, including mesh, point cloud, estimated trunk diameter, and branch count. The reference high-resolution point cloud scanned by a FARO TLS.}
    \label{fig_1}
\end{figure}

Previous research has investigated two primary strategies to address limited viewing angles for plant 3D reconstruction in the field. One strategy involved developing custom over-the-tree sensing platforms capable of capturing both sides of plants in a single pass \cite{Gongal2016, Sun2020, Williams2024}. Although this strategy offered dense multi-view coverage, it depends on highly specialized and costly equipment, making it less feasible for diverse agricultural fields. Another strategy scanned each side of plants separately and registered the resulting point clouds to reconstruct the complete structure \cite{Häni2020}. However, this strategy requires careful post-processing methods and is unreliable for scenes with less features or complex structure for point cloud registration such as apple orchards during the dormant season.

We identified that the core limitation of 2D imaging systems for 3D tree crop reconstruction lies in their dependence on expensive, purpose-built hardware. This requirement also demands highly customized software solutions, which ultimately restricts generalization capabilities. To bridge this gap, our study introduced a comprehensive approach that combines a data collection robot, a straightforward scanning strategy, and a novel robotic perception framework (Fig. \ref{fig_1}). This approach was designed to reduce hardware complexity and minimize post-processing requirements, offering a cost-effective, adaptable, and scalable solution for high-fidelity 3D apple tree reconstruction. We referred to the robot as AATBot (\textbf{A}miga-based \textbf{A}pple \textbf{T}ree Robot) and the robotic perception framework as DATR (\textbf{D}iffusion-based \textbf{A}pple \textbf{T}ree \textbf{R}econstruction) framework, respectively. 

The AATBot features a robust yet simplified mechanical design that minimizes vibration and ensures high-quality raw data, minimal hardware integration for quick assembly and easy adaptation, and flexible sensor adjustment for potential expansion (Fig. \ref{fig_1}). For data collection, we chose to operate the AATBot along a single side of the planting rows. This scanning strategy increases operational efficiency while introducing challenges related to limited viewpoints. To address the limited view challenge caused by this scan strategy, we incorporated strong 3D priors from diffusion models and large reconstruction model (LRM) in the DATR framework. This framework consists of two primary components: a background removal module and a reconstruction module (Fig. \ref{fig_3}). Both are intended to automate the reconstruction process and improve output quality for potential large-scale applications. The background removal module integrates onboard depth data with estimated depth maps from foundational monocular depth models to produce accurate tree masks from complex in-the-wild imagery. This significantly reduces the need for manual annotation. These foreground masks are then used to create object-centric inputs enriched with multi-modal cues such as color, semantic, and geometric information. These inputs are passed to the reconstruction module, which utilizes the generative capabilities of diffusion models and the strong spatial priors of LRM to infer detailed 3D tree geometry from sparse field imagery. Our major contributions in this study include: 

\begin{itemize}
\item We developed AATBot for efficient data collection and the DATR framework for accurate single-image-to-3D reconstruction in orchards, addressing domain-specific challenges inherent to these settings.

\item We enhanced diffusion model performance through multi-modal field data integration and demonstrated that synthetic 3D trees generated via a Real2Sim pipeline effectively train both the diffusion model and LRM, improving reconstruction quality.

\item The DATR framework achieved high geometric fidelity in zero-shot real-world apple tree reconstruction, enabling precise phenotypic trait measurement and supporting applications including yield estimation and digital twin development for crop management.
\end{itemize}

\section{Related Work}

Traditional 3D reconstruction methods, including structure-from-motion (SfM)\cite{snavely2006photo} and multi-view stereo (MVS)\cite{furukawa2010accurate}, estimate scene geometry through correspondence triangulation across calibrated images. While effective in controlled environments, these approaches struggle with sparse viewpoints, occlusions, and textureless surfaces due to their reliance on explicit feature matching. Neural radiance fields (NeRF)\cite{mildenhall2020nerf, zhang2020nerf++, barron2021mip} address these limitations through implicit volumetric scene representation, enabling photorealistic novel view synthesis and explicit 3D geometry extraction. However, NeRF methods still require dense multi-view images and accurate camera poses, constraining practical deployment. While few-shot NeRF variants \cite{Yu2021, Chen2021} leverage learned priors to reduce view requirements, they remain limited to specific object categories or struggle with complex real-world scenes. Recent advances combine diffusion models \cite{Ho2020} with score distillation sampling (SDS) \cite{Poole2022} or LRM \cite{Hong2023} to enable high-fidelity reconstruction from sparse inputs, generating plausible structures and fine details despite limited viewing conditions.

Sparse-view 3D reconstruction presents fundamental challenges due to geometric ambiguity and insufficient constraints. Early approaches utilized category-specific 3D templates as shape priors \cite{Roth2016, Goel2020, Gene-Mola2021, Marks2022}, achieving reasonable reconstructions within predefined object classes. While category-agnostic methods \cite{Niemeyer2020, Yariv2021} demonstrated broader generalization, they often failed to capture fine-grained details \cite{Yu2021}. Recent developments leverage diffusion models for multi-view guidance in image-to-3D reconstruction \cite{Liu2023, Shi2023, Qian2023, Liu2024}. Zero123 \cite{Liu2023} fine-tunes stable diffusion to synthesize novel views conditioned on input images and camera poses, subsequently distilling this knowledge to NeRF via SDS. However, this approach exhibits multi-view inconsistency and the "Janus" problem—where conflicting supervision signals produce geometries with multiple inconsistent faces. InstantMesh \cite{Xu2024} addresses these limitations using Zero123++ \cite{Shi2023} for multi-view consistent generation, followed by LRM-based reconstruction through triplane decoders \cite{Chan2022}. The Large Multi-view Gaussian (LGM) model \cite{Tang2024} further advances this paradigm by combining multi-view diffusion models \cite{Wang2023, Shi2023-mvdream} with 3D-aware U-Net architectures to directly predict high-resolution Gaussian fields, achieving efficient photorealistic 3D content creation with robust view consistency.

Agricultural applications increasingly adopt 3D reconstruction for yield estimation \cite{Häni2020, Gene-Mola2021}, crop monitoring \cite{Marks2022, Luo2023}, and crop modeling \cite{Zarei2024, Qiu2024, Qiu2025}. The field has progressed from traditional SfM and MVS to neural field methods \cite{Qiu2023, Smitt2023, Meyer2024, Jiang2025}, enabling continuous high-fidelity scene representations. Diffusion models have recently emerged in agricultural contexts for their capacity to generate detailed representations from limited or noisy inputs \cite{Yang2024, Kim2025}. \cite{Wang2024} employed diffusion-based frameworks to reconstruct detailed leaf geometry from single RGB images in controlled settings. Tree-D Fusion \cite{Lee2024} reconstructs 3D tree models from internet-sourced street-view images, prioritizing graphical realism. In contrast, our study developed a robotic perception framework for single-image-to-3D reconstruction of apple trees in field conditions, specifically targeting digital twin development for agricultural robotics. 

\section{PRELIMINARIES}

Diffusion models \cite{Ho2020} generate data by learning to reverse a noise-adding process, iteratively denoising Gaussian noise to produce realistic samples. Given an input $\mathbf{x_0}$, the forward process adds noise as:

\begin{equation}
\label{eq_diffusion_foward}
q(\mathbf{x}_t \mid \mathbf{x}_0) = \mathcal{N}(\mathbf{x}_t; \sqrt{\bar{\alpha}_t} \mathbf{x}_0, (1 - \bar{\alpha}_t) \mathbf{I}),
\end{equation}

where $\bar{\alpha_t}$ represents cumulative noise schedule parameters. The denoising network $\boldsymbol{\epsilon_\theta}$ is trained to predict the added noise by minimizing:

\begin{equation}
\label{eq_diffusion_loss}
\mathcal{L} = \mathbb{E}_{\mathbf{x}_0, \boldsymbol{\epsilon}, t} \left[ \left| \boldsymbol{\epsilon} - \boldsymbol{\epsilon}_\theta(\mathbf{x}_t, t) \right|^2 \right].
\end{equation}

Zero123 \cite{Liu2023} extends this framework for novel view synthesis by conditioning the denoising process on target camera viewpoints $\mathbf{v}$. The network predicts noise conditioned on both appearance and pose:

\begin{equation}
\label{eq_diffusion_loss_condition}
\boldsymbol{\epsilon}_\theta(\mathbf{x}_t, t \mid \mathbf{x}_0, \mathbf{v}),
\end{equation}

where $\mathbf{v}$ is encoded as learnable pose embeddings fused with image features. This enables 3D-aware content generation from single images by leveraging geometric priors without explicit multi-view supervision. Zero123++ \cite{Shi2023} enhances multi-view consistency through local and global conditioning mechanisms, modeling the joint distribution of six spatially coherent views in a 3×2 tile layout to ensure geometric consistency across viewpoints.

\section{Methods}

\subsection{Robot Platform}

Our AATBot is built on the Amiga mobile platform \cite{Farm-ng}, equipped with an NVIDIA Jetson AGX Orin Developer Kit to enable high-performance edge computing (Fig. \ref{fig_1}). A ZED X Mini stereo camera, featuring a 4-millimeter baseline and an integrated polarizing filter, is mounted on a centrally positioned vertical bar that allows for flexible camera alignment and future expansion with additional cameras. Two mechanically reinforced side bars provide structural support to stabilize the system and reduce the effects of uneven terrain during operation. The system captures RGB and depth video streams at a resolution of 1920×1200 pixels and a frame rate of 15 fps, using the ZED capture card for data transmission.

\begin{figure*}[t]
    \includegraphics[width=\linewidth]{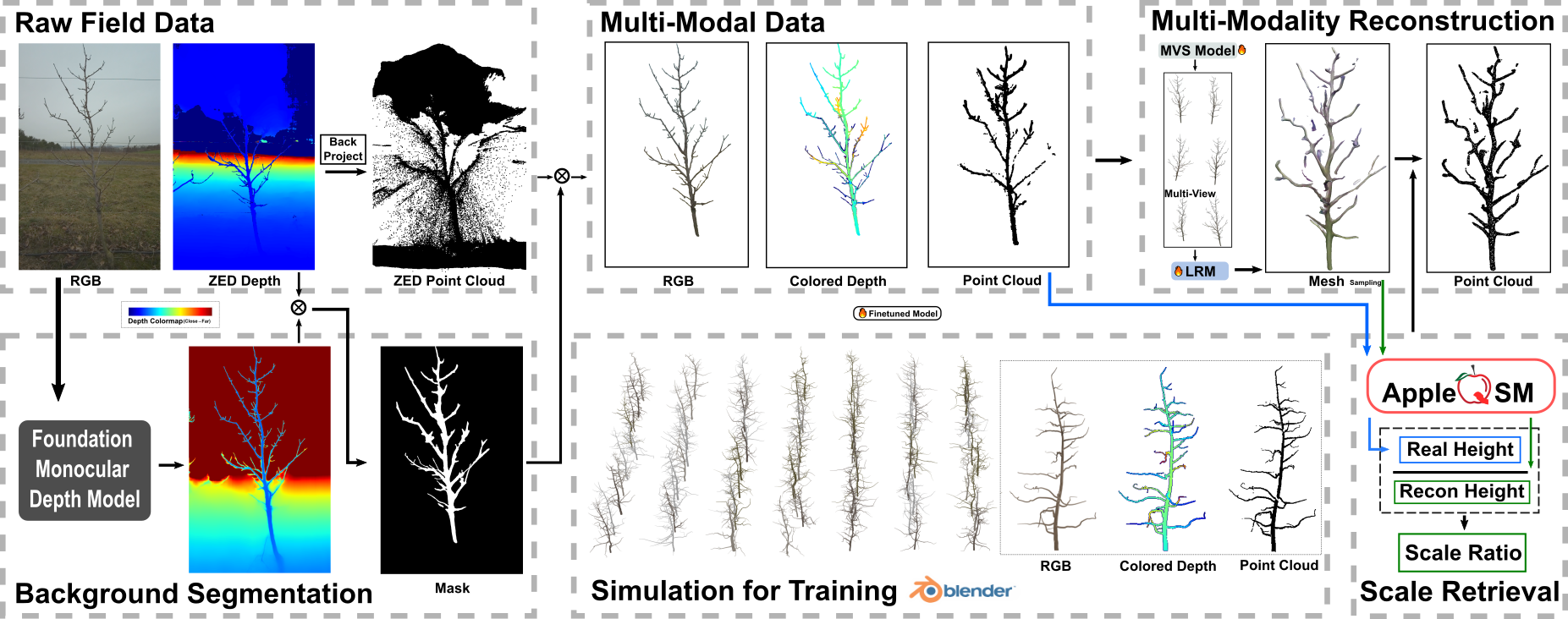}
    \caption{Overview of our DART reconstruction framework. The framework processes multi-modal field data through background segmentation to extract object-centric RGB, depth, and point cloud representations. These inputs feed into MM-Recon module for novel view synthesis, followed by LRM regression to generate 3D NeRF representations and extract meshes. Point clouds sampled from the meshes are scaled using geometric priors computed via AppleQSM. The MM-Recon models train exclusively on synthetic data generated via a Real2Sim pipeline, enabling zero-shot deployment in field conditions.}
    \label{fig_3}
\end{figure*}

\subsection{Robot Perception}

The DART reconstruction framework is composed of two core components: a background removal module that segments trees from complex field scenes and a reconstruction module for high-fidelity 3D reconstruction from single view (Fig. \ref{fig_3}).

\subsubsection{Background Removal} This module begins with a distance-based filtering procedure using metric depth from the ZED stereo camera to effectively eliminate far-distanced background elements. However, the ZED depth estimates become unreliable in regions with minimal disparity, especially around the sky, resulting in poor segmentation of fine structures such as branches with the sky background. To address this, we incorporate depth predictions from a monocular depth estimation model (DepthPro \cite{Bochkovskii2024}) to construct a sky mask that accurately filters out sky pixels. Additionally, we generate a ground mask using a Z-filtering technique applied to the 3D point cloud reconstructed via unprojection of the ZED depth. This removes pixels corresponding to near-ground artifacts. Finally, to eliminate residual interference from adjacent trees, we apply a K-means clustering algorithm to segment and discard spatially inconsistent clusters in the image space.

\subsubsection{Multi-Modality Reconstruction}
This module (\textbf{MM-Recon}) integrates a view-conditioned diffusion model for multi-view synthesis with an LRM for neural field regression. Unlike conventional view-conditioned models that rely solely on RGB input, ours leverages multi-modal inputs: RGB images, colorized depth maps encoding semantic cues, and partial point clouds capturing geometric structure. This multi-modal conditioning enhances both fidelity and spatial consistency of synthesized views in complex natural scenes. While CLIP encodes RGB and colorized depth maps, we employ a transformer-based encoder for point cloud data similar to \cite{Huang2025}:

\begin{equation}
\label{eq_mmodality}
\mathbf{z}_\text{rgb}, \mathbf{z}_\text{depth} = \text{CLIP}(\text{RGB}, \text{D}), \quad
\mathbf{z}_\text{pc} = \text{Transformer}_\text{PC}(P)
\end{equation}
\begin{equation}
\label{eq_diffusion_fusion}
\mathbf{z}_\text{fused} = \text{Concat}(\mathbf{z}_\text{rgb}, \mathbf{z}_\text{depth}, \mathbf{z}_\text{pc})
\end{equation}

where RGB, D, and P denote the input RGB image, colorized depth map, and point cloud, respectively. The concatenated embeddings form the multi-modal context, enabling joint reasoning over visual and geometric cues for accurate novel view synthesis.

Both models are trained on synthetic datasets generated via a Real2Sim pipeline \cite{Qiu2025}, which produces diverse 3D apple tree models with high geometric fidelity and controllable structural variation. This approach provides high-quality ground truth supervision while facilitating Sim2Real generalization.

\subsubsection{Scale Retrieval} A fundamental challenge in image-based 3D reconstruction especially under limited-view conditions is the inherent scale ambiguity, where the reconstructed geometry lacks a grounded metric reference. To address this, we introduced a scale retrieval mechanism that leverages geometric priors derived from the ZED point cloud. Specifically, we compute the ratio of tree height between a DATR reconstructed tree and ZED depth unprojected one (Eq. \ref{eq_scale}). 

\begin{equation}
\label{eq_scale}
s = \frac{H_\text{ZED}}{H_\text{rec}}
\end{equation}

\( H_\text{ZED} \) denotes the estimated tree height from the ZED point cloud, and \( H_\text{rec} \) is the corresponding height of the reconstructed model. The scale factor \( s \) is applied uniformly to the reconstruction to align it with real-world metric space.

Through empirical evaluation, we found that tree height serves as a more stable and reliable physical reference for scale calibration compared to trunk diameter, which is highly sensitive to incomplete observations and noise. By aligning the reconstructed tree to this measured height, we effectively anchor the reconstruction to a real-world metric scale, improving both quantitative accuracy and practical applicability for downstream phenotyping and robotic tasks.

\section{Experiments}

\subsection{Dataset}

To support both training and evaluation of our proposed framework, we curated a combination of real-world and synthetic datasets designed to capture a diverse range of apple tree structures and viewpoints.

\subsubsection{Field Dataset} This dataset was collected in New York State, USA, during the offseason when trees were dormant. For evaluation only, we selected six apple trees exhibiting diverse branching patterns and structural complexity. Data acquisition was performed using our AATBot platform, which traversed along planting rows at a constant speed of 2 mile per hour in the cruise, resulting in around 30 seconds to acquire 6 trees.

\subsubsection{Real2Sim Dataset} 
\begin{figure}[!t]
    \includegraphics[width=\linewidth]{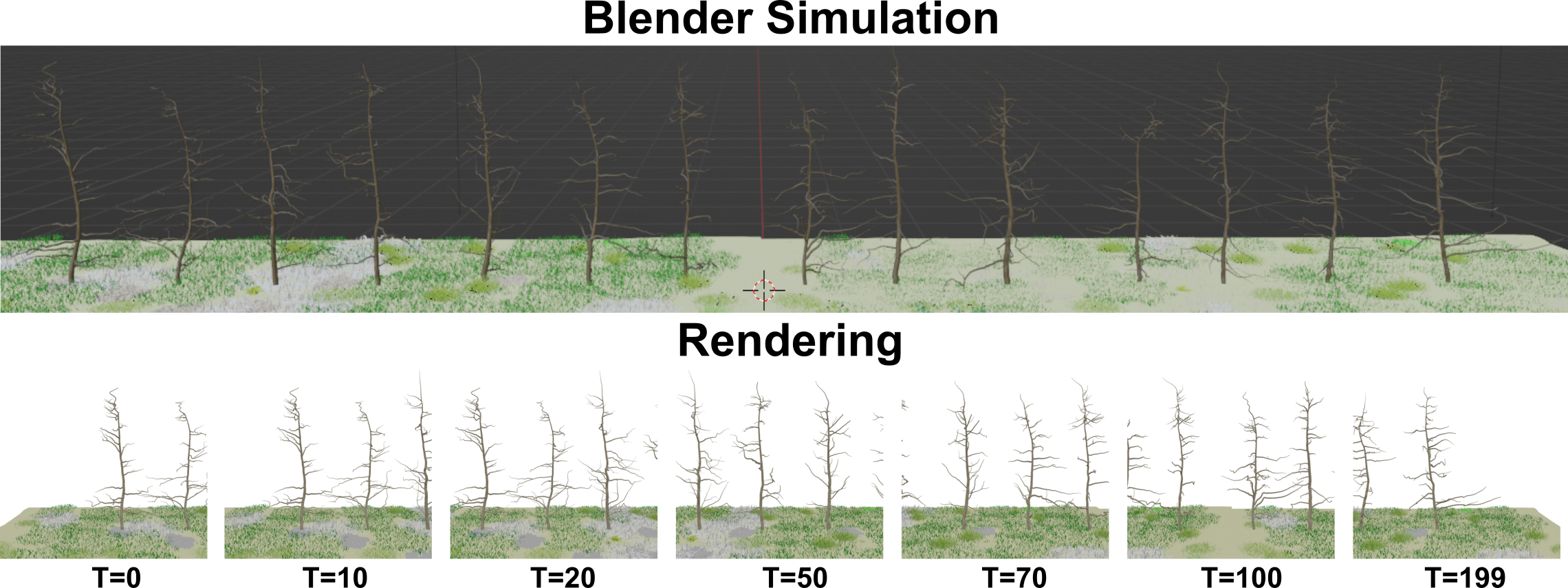}
    \caption{Synthetic trees from R2S-Val in Blender. Bottom: R2S-NeRF dataset creation showing rendered sequences from a virtual camera simulating AATBot's field trajectory.}
    \label{fig_4}
\end{figure}

Our synthetic dataset comprises 5,000 procedurally generated apple tree models created using a Real2Sim generator, designed to exhibit diverse realistic geometries and topologies (Fig. \ref{fig_4} Top). A total of 30 trees was reserved for validation (R2S-Val) while the remaining 4970 trees were for training. Three datasets were generated for different training purposes: (1) R2S-Diffusion: Following \cite{Shi2023}, we rendered 4,970 trees with one query image and six target views at fixed elevation and azimuth offsets for diffusion model fine-tuning; (2) R2S-LRM: Following \cite{Hong2023}, we generated 32 randomly distributed views per tree for the same 4,970 trees for LRM fine-tuning; and (3) R2S-NeRF: Using a simulated AATBot trajectory, we rendered the trees in R2S-Val with 15-30 views each, where a virtual camera with ZED-equivalent parameters traverses at 2 mph along planting rows (Fig. \ref{fig_4} Bottom).

\subsection{Experiment Setup}

We evaluated reconstruction performance using diverse baselines spanning learning-based, sensor-based, and traditional geometric approaches. The ground truth comprised mesh sampled point clouds from R2S-Val trees for synthetic evaluation and TLS scans (FARO S350) of 6 field trees captured from 6 viewpoints (3 per side along crop rows) for approximately 3 hours, providing robust benchmarks for both domains.

For learning-based methods, we selected Nerfacto \cite{Nerfstudio} representing neural field approaches and InstantMesh \cite{Xu2024} as state-of-the-art diffusion-based reconstruction. In addition to these two, we included a finetuned variant (InstantMesh-FT) for comparison with our MM-Recon models on R2S-Val. For field evaluation only, we included two more non-learning baselines: a sensor-based pipeline reconstructing directly from ZED RGB-D streams, and traditional COLMAP \cite{COLMAP}. 

\subsection{Model Training}

Nerfacto performed per-scene optimization using R2S-NeRF with ground-truth poses for synthetic evaluation and COLMAP-estimated poses from field sequences for real-world testing. Both InstantMesh-FT and MM-Recon models were fine-tuned exclusively on synthetic data through two-stage training: Zero123++ on R2S-Diffusion followed by LRM on R2S-LRM, each trained for 50 epochs with learning rate 1e-5. The key distinction lies in their input modalities: our MM-Recon leverages multi-modal inputs (RGB, colorized depth, and point cloud) with the point cloud transformer ($\text{Transformer}_\text{PC}$) trained from scratch jointly with Zero123++, while InstantMesh-FT processes only RGB images using the standard Zero123++ architecture.

\subsection{Evaluation Metrics}

We evaluated reconstruction results using geometric metrics for structural accuracy and domain-specific traits for agriculture-specific relevance and performance.

Geometric evaluation was conducted by comparing reconstructed point clouds to ground-truth references using Chamfer Distance (CD-$l_2$) and Jensen-Shannon divergence (JSD). The Chamfer Distance quantifies local geometric completeness and point-level fidelity, while the JSD evaluates the similarity of global point distributions. These metrics were computed after aligning the reconstructed and ground-truth point clouds using iterative closest point with the criterion of $1e^{-7}$ RMS difference. Due to the lack of complete geometric ground truth in field conditions, these geometric evaluations were restricted to the synthetic dataset.


Domain-specific evaluation focused on traits critical for horticultural applications, such as trunk diameter and branch count. These metrics were selected for their importance in plant phenotyping and production management. Ground-truth measurements for these traits were obtained manually in the field. For the reconstructed models, trait estimates were extracted using AppleQSM \cite{Qiu2024-AppleQSM}, a widely used tree structure analysis tool. This evaluation was conducted on the field dataset and applied to all models capable of producing scale-consistent reconstructions, including InstantMesh, Nerfacto, the sensor-based ZED pipeline, the COLMAP-based photogrammetry method, and our MM-Recon module.

\begin{figure}[!t]
    \includegraphics[width=\linewidth]{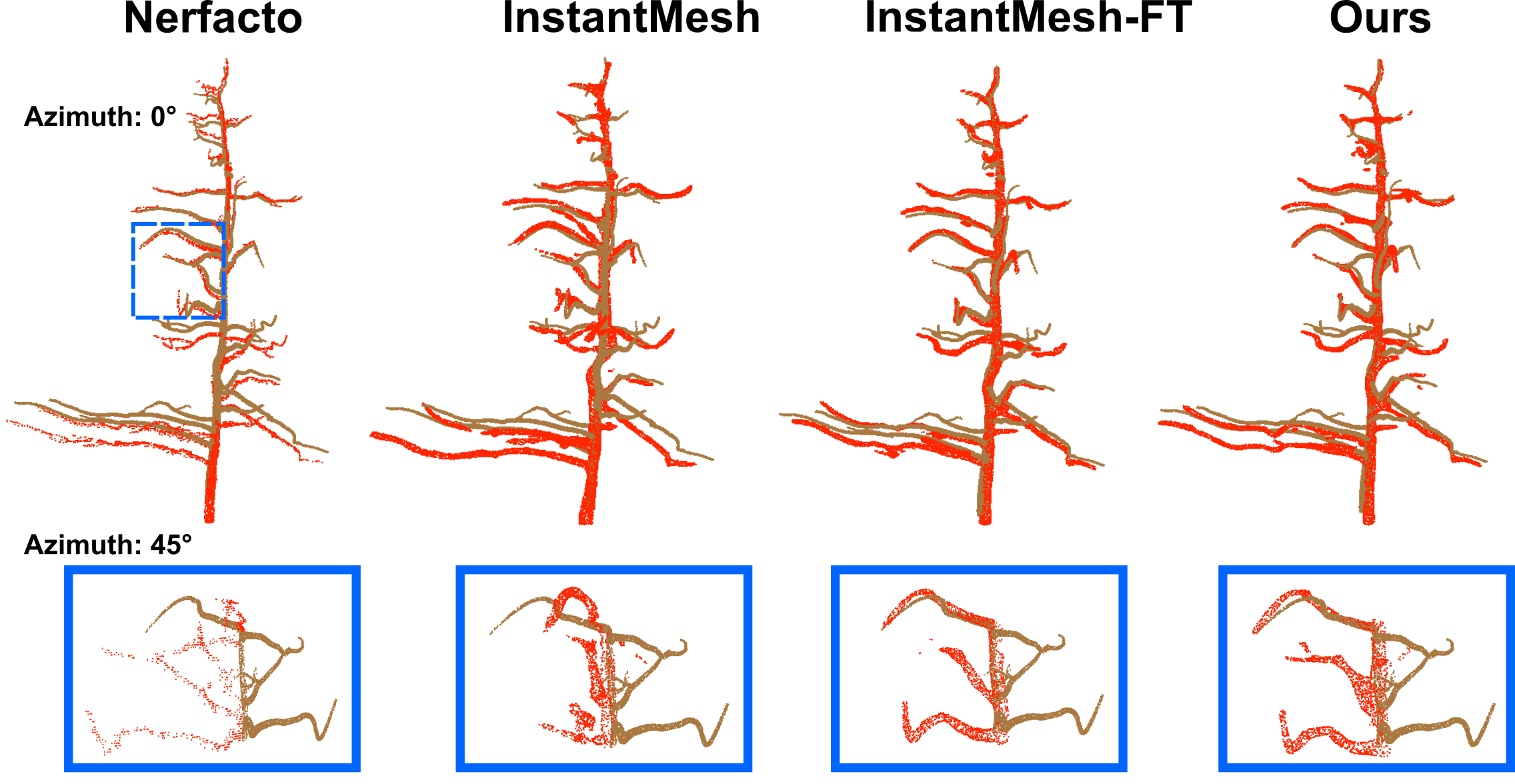}
    \caption{Qualitative comparison of reconstructed tree structures from different methods on the synthetic dataset. Reconstructed point cloud (Red) was registered to the GT point cloud using ICP. Tree-level views are shown from an azimuth angle of 0°, while the zoomed-in regions (blue boxes) are captured from an azimuth of 45° to highlight structural discrepancies. The zoom-in reveals reconstruction errors that appear correct when viewed only from the 0° perspective.}
    \label{fig_5}
\end{figure}

\section{Results}

\begin{table}[!b]
    \caption{Geometric Reconstruction Quality Metrics (Synthetic Trees).}
    \label{tab:reconstruction_metrics}
    \centering
    \begin{tabular}{|l|cc|cc|}
    \hline
    \multirow{2}{*}{\textbf{Method}} & \multicolumn{2}{c|}{\textbf{CD-$l_2$ (m)}} & \multicolumn{2}{c|}{\textbf{JSD}} \\
    \cline{2-5}
     & mean & std & mean & std \\
    \hline
    Nerfacto         & 0.0514 & 0.0185 & 0.6643 & 0.0220 \\
    InstantMesh      & 0.0355 & 0.0136 & 0.6644 & 0.0258 \\
    InstantMesh-FT   & 0.0252 & 0.0088 & 0.6559 & 0.0328 \\
    Ours             & 0.0246 & 0.0095 & 0.6559 & 0.0300 \\
    \hline
    \end{tabular}
\end{table}

\subsection{Synthetic Data Evaluation}

All methods reconstructed the overall tree geometry reasonably well, capturing trunks and major branches under sparse-view conditions (Fig. \ref{fig_5}). However, sparse views significantly hindered the accurate recovery of fine-scale topology, such as bifurcations and subtle curvature, which are more sensitive to viewpoint occlusion and depth ambiguity. These limitations became evident when comparing zoom-in views captured from different azimuth angles (Fig. \ref{fig_5} Bottom). Among the methods evaluated, Nerfacto was affected the most by sparse inputs and produced visibly incomplete and fragmented reconstructions.

These challenges are more evidently reflected in the quantitative evaluation (Tab. \ref{tab:reconstruction_metrics}). Nerfacto exhibited poor performance in synthetic settings, with high CD and JSD values reflecting its dependence on dense views and textured backgrounds, both absent in our planar-view synthetic data. Diffusion-based methods showed marked improvements: InstantMesh reduced CD but maintained high JSD, indicating better local geometry but limited global distribution. Fine-tuning on our synthetic dataset (InstantMesh-FT) yielded lower CD and modest JSD reduction, demonstrating adaptation to tree-specific structural priors. Our MM-Recon module achieved the lowest CD through multi-modal feature integration, though consistently high JSD across all methods reveals a fundamental challenge in global spatial understanding from sparse views.

Domain-specific evaluation on trunk diameter and branch count further quantitatively reinforced the superiority of our method in accurately reconstructing biologically relevant tree attributes (Tab. \ref{tab:trunk_diameter_lpy} and \ref{tab:branch_count_lpy}). For trunk diameter estimation, our method reduced the mean MAE to 0.47 cm and MAPE to 8.67\%, a substantial improvement over InstantMesh-FT (0.82 cm, 14.97\%) and a dramatic reduction compared to Nerfacto and InstantMesh, both of which exceed 2 cm and 40\% MAPE. Similar trends are observed in the branch count evaluation, where our method achieves a mean MAE of 4.13 and a MAPE of 17.18\%, significantly outperforming both InstantMesh (8.53 MAE, 35.78\% MAPE) and Nerfacto (8.13 MAE, 34.25\% MAPE). The reduced errors in trunk diameter and branch count indicate not only improved mesh fidelity but also better preservation of biologically meaningful structural details.

\begin{table}[!t]
    \begin{threeparttable}
    \caption{Quantitative Trunk Diameter Evaluation (Synthetic Trees).}
    \label{tab:trunk_diameter_lpy}
    \centering
    \setlength{\tabcolsep}{3pt}
    \begin{tabular}{|p{2cm}|p{0.9cm}p{0.9cm}p{0.9cm}|p{0.9cm}p{0.9cm}p{0.9cm}|}
    \hline
    \textbf{Method} & \multicolumn{3}{c|}{\textbf{MAE (cm)}} & \multicolumn{3}{c|}{\textbf{MAPE (\%)}} \\
    \cline{2-7}
     & \textbf{mean} & \textbf{std} & \textbf{75th} & \textbf{mean} & \textbf{std} & \textbf{75th} \\
    \hline
    R2S-Val         & 0.09 & 0.04 & 0.12 & 1.64  & 0.61  & 2.05 \\
    Nerfacto         & 3.10 & 5.57 & 6.85 & 61.71 & 123.23 & 144.76 \\
    InstantMesh      & 2.24 & 1.53 & 3.27 & 40.93 & 27.61 & 59.54 \\
    InstantMesh-FT   & 0.82 & 0.33 & 1.04 & 14.97 & 5.80  & 18.88 \\
    Ours             & 0.47 & 0.26 & 0.64 & 8.67  & 4.79  & 11.90 \\
    \hline
    \end{tabular}
    \begin{tablenotes}
    \footnotesize
    \item R2S-Val represents GT point cloud processed through AppleQSM to establish baseline measurement variability.
    \end{tablenotes}
    \end{threeparttable}
\end{table}

\begin{table}[!t]
\caption{Quantitative Branch Count Evaluation (Synthetic Trees).\label{tab:branch_count_lpy}}
\centering
\setlength{\tabcolsep}{3pt}
\begin{tabular}{|p{2cm}|p{0.9cm}p{0.9cm}p{0.9cm}|p{0.9cm}p{0.9cm}p{0.9cm}|}
\hline
\textbf{Method} & \multicolumn{3}{c|}{\textbf{MAE (count)}} & \multicolumn{3}{c|}{\textbf{MAPE (\%)}} \\
\cline{2-7}
 & \textbf{mean} & \textbf{std} & \textbf{75th} & \textbf{mean} & \textbf{std} & \textbf{75th} \\
\hline
R2S-Val         & 1.47 & 1.86 & 2.72 & 6.06  & 7.07  & 10.82 \\
Nerfacto         & 8.13 & 4.76 & 11.34 & 34.25 & 18.43 & 46.67 \\
InstantMesh      & 8.53 & 3.46 & 10.87 & 35.78 & 13.76 & 45.05 \\
InstantMesh-FT   & 5.70 & 2.51 & 7.39  & 23.85 & 9.58  & 30.31 \\
Ours             & 4.13 & 2.23 & 5.64  & 17.18 & 8.66  & 23.02 \\
\hline
\end{tabular}
\end{table}

\begin{figure*}[!t]
    \includegraphics[width=\linewidth]{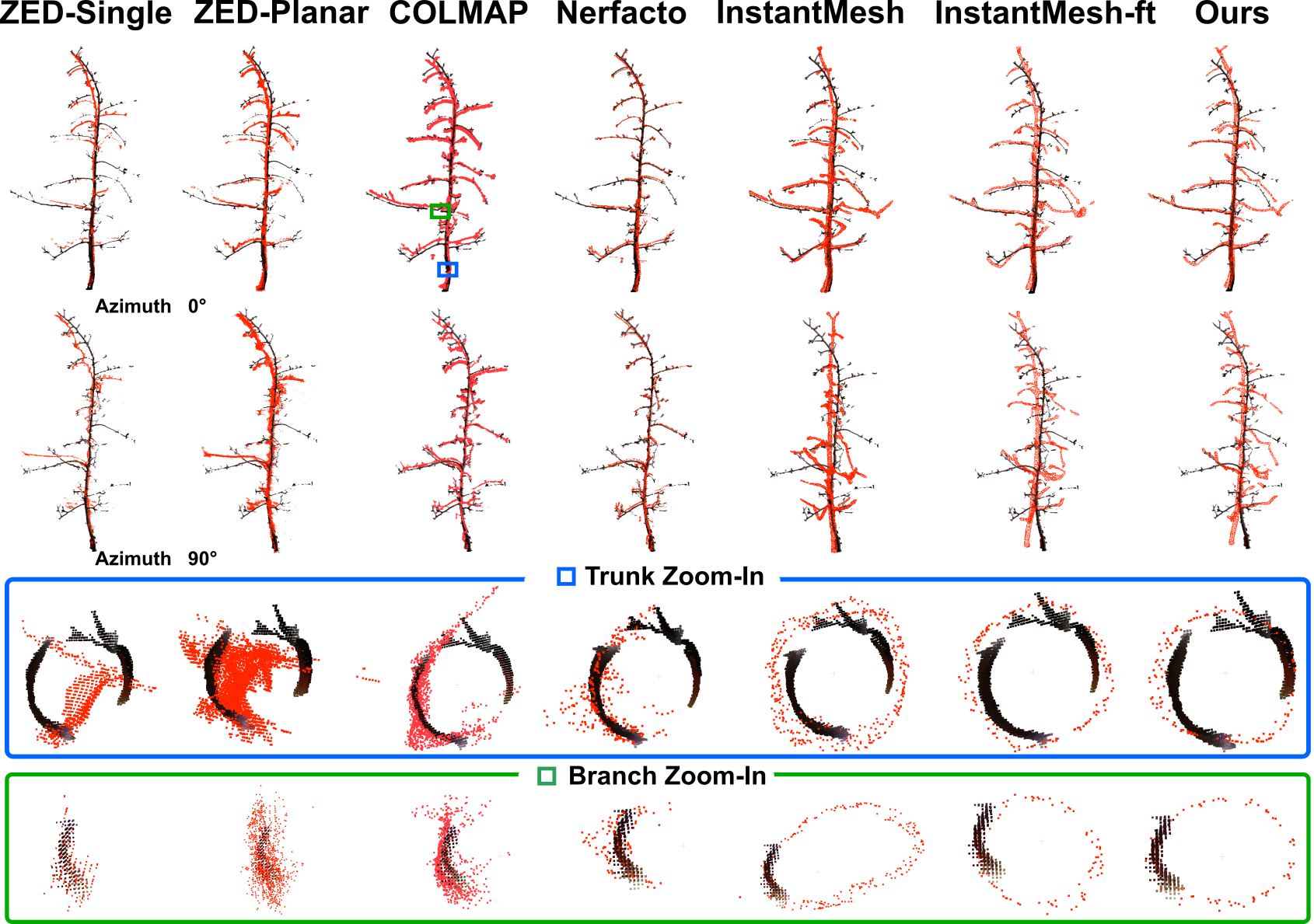}
    \caption{Qualitative comparison of reconstructed tree structures from different methods with a focus on trunk and branch geometrical fidelity. Reconstructed point cloud (Red) was registered to the FARO point cloud using ICP. The top two row presents full-tree visualizations from two views where the first row is the in-field camera view. The bottom two rows show zoomed-in cross-sectional views of the trunk and a primary branch, respectively. The cross-sectional point clouds were cropped and registered again on the cropped point cloud.}
    \label{fig_7}
\end{figure*}

\subsection{Field Data Evaluation}

Qualitative results highlighted two significant trends across the methods (Fig. \ref{fig_7}). First, although the global coarse alignment is consistently achieved across all methods, capturing the overall shape and orientation of the tree, the precise fine-grained structural alignment remains challenging. Second, only diffusion-based methods demonstrated the capability to infer relatively complete and plausible geometries under sparse-view conditions. This is crucial from a downstream phenotyping standpoint, as supported by the quantitative evaluation shown (Tab. \ref{tab:mae_mape_trunk} and \ref{tab:mae_mape_branch}). Our method achieved a mean trunk diameter MAE of 0.68 cm and a branch count MAE of 3.67, which is comparable to the FARO point cloud, while enhancing data collection throughput by nearly 360 times.

\subsubsection{ZED-based methods} ZED-Single and ZED-Planar introduced significant noise and artifacts, particularly in fine-structured regions like branches and trunk perimeters. This noise leads to high variability and unreliable geometry, hindering the accurate trait estimation. As a result, \textbf{trunk diameter and branch count could not be evaluated} for either ZED method, which violate the assumptions of structural analysis tools like AppleQSM.

\subsubsection{COLMAP and Nerfacto} COLMAP reconstructed globally coherent and relatively dense point clouds, demonstrating better resilience to sparse-view inputs than Nerfacto. Notably, it achieved the most accurate branch count estimation (mean MAE=2.83). This likely reflects its ability to preserve continuous geometry at trunk-branch junctions -- a critical factor for counting. However, COLMAP still produced noisier geometry than learning-based models, particularly in low-texture regions, contributing to a moderately higher trunk diameter error.

\subsubsection{Diffusion-based methods} Diffusion-based methods, evaluated in a zero-shot setting after training on synthetic data, revealed varying levels of cross-domain generalization. InstantMesh produced coherent and relatively complete reconstructions, yet consistently overestimated trunk and branch thickness. This is reflected in a trunk diameter MAE of 1.44 cm and a branch count MAE of 6.33, suggesting a domain gap between training and real-world scenes. Fine-tuning with synthetic apple tree data (InstantMesh-FT) significantly improved trait accuracy, lowering the trunk diameter MAE to 0.79 cm and branch count MAE to 4.67, thereby reducing both bias and variance. This underscores the utility of domain-specific datasets in improving performance. Our method achieved the best overall accuracy with a trunk diameter MAE of 0.68 cm and a branch count MAE of 3.67. The improvement stems from its multi-modal conditioning on RGB, depth, and point cloud data, and the use of a transformer-based encoder that enhances structural reasoning. The resulting reconstructions align closely with FARO ground truth and exhibit minimal noise even under sparse, real-world view conditions. All in all, point clouds genrated by our MM-Recon module showed comparable completeness and quality to those collected using a TLS scanner, suggesting its efficacy for creating high-fidelity 3D tree models in agricultural fields from the 3D computer vision perspective. 

Branch count remained as a non-trivial error margin for diffusion-based methods. A key contributor to this error is the misrepresentation or miss of branches that grow toward the camera (Fig. \ref{fig_7}). These branches are captured as highly foreshortened or sparsely sampled point sets, lying along the camera frustum’s depth axis. In such configurations, even diffusion-based models struggle to infer accurate geometry and orientation due to the lack of spatial cues and view diversity. A promising direction to address this limitation is to incorporate planar views into the diffusion framework, leveraging the more complete azimuth coverage.

\begin{table}[!t]
\begin{threeparttable}
\caption{Quantitative Trunk Diameter Evaluation (Field Trees).}
\label{tab:mae_mape_trunk}
\centering
\setlength{\tabcolsep}{3pt}
\begin{tabular}{|p{2cm}|p{0.9cm}p{0.9cm}p{0.9cm}|p{0.9cm}p{0.9cm}p{0.9cm}|}
\hline
\textbf{Method} & \multicolumn{3}{c|}{\textbf{MAE (cm)}} & \multicolumn{3}{c|}{\textbf{MAPE (\%)}} \\
\cline{2-7}
 & \textbf{mean} & \textbf{std} & \textbf{75th} & \textbf{mean} & \textbf{std} & \textbf{75th} \\
\hline
FARO             & 0.66 & 0.26 & 0.83 & 13.86 & 5.16 & 17.33 \\
ZED-Single       & --   & --   & --   & --    & --    & --    \\
ZED-Planar       & --   & --   & --   & --    & --    & --    \\
COLMAP           & 1.71 & 0.65 & 2.14 & 36.41 & 14.60 & 46.25 \\
Nerfacto         & 1.11 & 0.59 & 1.51 & 22.87 & 10.13 & 29.69 \\
InstantMesh      & 1.44 & 1.00 & 2.12 & 30.56 & 22.40 & 45.66 \\
InstantMesh-FT   & 0.79 & 0.60 & 1.20 & 15.93 & 11.63 & 23.77 \\
Ours             & 0.68 & 0.34 & 0.91 & 13.77 & 5.89  & 17.74 \\
\hline
\end{tabular}
\begin{tablenotes}
\footnotesize
\item ZED-based methods produced excessive noise and gaps that violated AppleQSM's geometric assumptions, preventing trait extraction.
\end{tablenotes}
\end{threeparttable}
\end{table}

\begin{table}[!t]
\caption{Quantitative Branch Count Evaluation (Field Trees).\label{tab:mae_mape_branch}}
\centering
\setlength{\tabcolsep}{3pt}
\begin{tabular}{|p{2cm}|p{0.9cm}p{0.9cm}p{0.9cm}|p{0.9cm}p{0.9cm}p{0.9cm}|}
\hline
\textbf{Method} & \multicolumn{3}{c|}{\textbf{MAE (count)}} & \multicolumn{3}{c|}{\textbf{MAPE (\%)}} \\
\cline{2-7}
 & \textbf{mean} & \textbf{std} & \textbf{75th} & \textbf{mean} & \textbf{std} & \textbf{75th} \\
\hline
FARO           & 3.67 & 2.56 & 5.39 & 18.45 & 13.14 & 27.31 \\
ZED-Single     & --   & --   & --   & --    & --    & --    \\
ZED-Planar     & --   & --   & --   & --    & --    & --    \\
COLMAP         & 2.83 & 3.02 & 4.87 & 14.34 & 12.95 & 23.06 \\
nerfacto       & 6.17 & 5.79 & 10.07 & 29.93 & 26.52 & 47.81 \\
InstantMesh      & 6.33 & 4.64 & 9.46 & 26.67 & 16.49 & 37.78 \\
InstantMesh-FT   & 4.67 & 2.49 & 6.35 & 22.06 & 9.17  & 28.24 \\
Ours             & 3.67 & 3.09 & 5.75 & 15.40 & 12.33 & 23.71 \\
\hline
\end{tabular}
\end{table}


\section{Conclusion}

This work introduced DATR, a comprehensive reconstruction framework that addresses single-image-to-3D reconstruction challenges in semi-structured orchard environments. The core innovation lies in our MM-Recon module, which integrates multi-modal field data including RGB images, colorized depth maps, and point clouds through specialized encoders, enabling robust structural reasoning from sparse viewpoints. Additionally, our scale retrieval module leverages geometric priors to automatically align reconstructed meshes to real-world dimensions, eliminating manual calibration while preserving biologically relevant traits. To enable training without costly field annotations, we developed a Real2Sim dataset comprising procedurally generated apple trees with diverse geometries and tailored rendering protocols. Quantitative and qualitative evaluations demonstrate that DATR achieves TLS-comparable accuracy while providing approximately 360× throughput improvement. These results validate the framework's strong Sim2Real generalization and practical potential for supporting downstream applications including automated phenotyping, yield estimation, and digital twin development for precision agriculture.

\section*{Acknowledgments}
The present study was supported by the USDA NIFA Hatch project (accession No. 1025032) and USDA NIFA Specialty Crop Research Initiative (award No. 2020-51181-32197). The authors would gratefully thank Kaspar Kuehn for helping field measurements.

\textbf{This work has been submitted to the IEEE for possible publication. Copyright may be transferred without notice, after which this version may no longer be accessible.}


 





\end{document}